\documentclass[3p]{elsarticle}

\usepackage{hyperref}

\usepackage{microtype}
\usepackage{graphicx}
\usepackage{subfigure}
\usepackage{booktabs} 
\usepackage{enumitem}

\usepackage{amsfonts}
\usepackage{calrsfs}
\DeclareMathAlphabet{\pazocal}{OMS}{zplm}{m}{n}
\usepackage{amsmath}

\newcommand{\eg}{{\em e.g.}}

\newcommand{\ie}{{\em i.e.}}
\newcommand{\etc}{{\em etc.}}
\newcommand{\RNum}[1]{\uppercase\expandafter{\romannumeral #1\relax}}

\DeclareMathAlphabet{\mathcal}{OMS}{cmsy}{m}{n}

\usepackage[dvipsnames]{xcolor}
\usepackage{float}

\journal{Applied Soft Computing}

\begin{document}

\begin{frontmatter}

\title{MO-PaDGAN: Reparameterizing Engineering Designs for Augmented Multi-Objective Optimization}


\author[weiaddress]{Wei Chen\corref{mycorrespondingauthor}}
\cortext[mycorrespondingauthor]{Corresponding author}
\ead{wei.wayne.chen@northwestern.edu}

\author[faezaddress]{Faez Ahmed}

\address[weiaddress]{Northwestern University, Evanston, IL 60208, USA}
\address[faezaddress]{Massachusetts Institute of Technology, Cambridge, MA 02139, USA}

\begin{abstract}
Multi-objective optimization is key to solving many Engineering Design problems, where design parameters are optimized for several performance indicators. However, optimization results are highly dependent on how the designs are parameterized. Researchers have shown that deep generative models can learn compact design representations, providing a new way of parameterizing designs to achieve faster convergence and improved optimization performance. Despite their success in capturing complex distributions, existing generative models face three challenges when used for design problems: 1)~generated designs have limited design space coverage, 2)~the generator ignores design performance, and 3)~the new parameterization is unable to represent designs beyond training data. To address these challenges, we propose MO-PaDGAN, which adds a Determinantal Point Processes based loss function to the generative adversarial network to simultaneously model diversity and (multi-variate) performance. MO-PaDGAN can thus improve the performances and coverage of generated designs, and even generate designs with performances exceeding those from training data. When using MO-PaDGAN as a new parameterization in multi-objective optimization, we can discover much better Pareto fronts even though the training data do not cover those Pareto fronts. In a real-world multi-objective airfoil design example, we demonstrate that MO-PaDGAN achieves, on average, an over 180\% improvement in the hypervolume indicator when compared to the vanilla GAN or other state-of-the-art parameterization methods.
\end{abstract}


\end{frontmatter}


\section{Introduction}

The goal of many Engineering Design applications is to find designs which perform well on multiple performance goals. For example, bike engineers want to design the geometry of road bikes which are lightweight, have less drag and exhibits good ride quality.
Such design applications often have problems with huge collections of data (CAD models, images, microstructures, \etc{}) with hundreds of features and multiple functionalities. These designs are commonly represented using parametric modeling, which defines design geometry using hard-coded features and constraints~\cite{chang2016design}. 
Once a parametric model is defined for a design, it can be used for design synthesis (generating new designs by changing design parameters) or for design optimization (finding optimal parameters which maximize the desirable performance goals). 
However, there are two issues with traditional hard-coded parametric modeling: 1)~the parameterization is often not flexible to cover all possible variations of the chosen type of designs, and 2)~the parametric representation may contain a large number of parameters, which often leads to infeasible designs and increased optimization cost. 
To alleviate these issues, researchers in Engineering Design have used machine learning techniques (deep generative models in particular) to replace hard-coded parametric models with compact representations learned from data~\cite{oh2019deep, burnap2019design, shu20203d}. Those learned representations can then be treated as parameterizations that are fed into a design optimization method to find one or more designs with optimal performance measures. Variational autoencoders (VAEs)~\cite{kingma2013auto} and Generative Adversarial Networks (GANs)~\cite{goodfellow2014generative} are the two most commonly used deep generative models for these tasks. These models are applied in design optimization over domains including microstructural design~\cite{yang2018microstructural}, 3D modeling~\cite{zhang20193d}, and aerodynamic shape design~\cite{chen2020airfoil}. However, existing generative models, whose goal is learning the distribution of existing designs, face three challenges when being used for parameterizing designs: 1)~generated designs may have limited design space coverage (\eg, when mode collapse happens with GANs), 2)~the generator ignores design performance, and 3)~the new parameterization is unable to represent designs outside the training data.

In this work, we address all the three challenges by modifying the architecture and loss of deep generative models, which allows simultaneous maximization of generated designs' diversity and (possibly multivariate) performance. With such a setting, we develop a new variant of GAN, named MO-PaDGAN (Multi-Objective Performance Augmented Diverse Generative Adversarial Network). 
This new model architecture is based on the PaDGAN architecture~\cite{chen2020padgan}, where both diversity and scalar performance are modeled by Determinantal Point Processes (DPPs)~\cite{kulesza2012determinantal}. 
 The current work extends the PaDGAN architecture~\cite{chen2020padgan} to account for multivariate performance and shows how the learned representations can be used for solving multi-objective optimization problems. 
We use MO-PaDGAN as a parameterization in multi-objective optimization tasks and demonstrate that it contributes to large improvements in the final Pareto optimal solutions when compared to a baseline GAN model and other state-of-the-art parameterizations. 
The contributions of this work are as follows:
\setlist{nolistsep}
\begin{enumerate}[noitemsep]
    \item We propose a new GAN architecture---MO-PaDGAN, which generalizes the ability to model a single performance measure in performance-augmented GANs ~\cite{chen2020padgan} to multiple performance measures.
    \item We show that the latent representation learned by MO-PaDGAN enhances performance and diversity in the learnt parameterization.
    \item We show that MO-PaDGAN parameterization leads to large improvements in multi-objective optimization solutions. For a real-world airfoil design example, the method leads to \emph{over 180\% average improvement} of the hypervolume indicator compared to three state-of-the-art methods.
    \item We demonstrate that MO-PaDGAN discovers novel high-performance designs that it had not seen from existing data.
    \item We demonstrate the generalizability of the proposed approach by showing large improvements in two synthetic and one real-world example. We also show the approach is optimization-method agnostic by reporting comparisons for both multi-objective Bayesian optimization and evolutionary algorithms.
\end{enumerate}

\section{Background}

Below we provide background on Generative Adversarial Networks and Determinantal Point Processes, which are the two key ingredients of our method.

\subsection{Generative Adversarial Nets}
\label{sec:gan}

Generative Adversarial Networks~\cite{goodfellow2014generative} model a game between a generative model (\textit{generator}) and a discriminative model (\textit{discriminator}). The generator $G$ maps an arbitrary noise distribution to the data distribution (\ie, the distribution of designs in our scenario), thus can generate new data; while the discriminator $D$ tries to perform classification, \ie, to distinguish between real and generated data. Both $G$ and $D$ are usually built with deep neural networks. As $D$ improves its classification ability, $G$ also improves its ability to generate data that fools $D$. 
Thus, a GAN has the following objective function:
\begin{equation}
\min_G\max_D V(D,G) = \mathbb{E}_{\mathbf{x}\sim P_{data}}[\log D(\mathbf{x})] + \mathbb{E}_{\mathbf{z}\sim P_{\mathbf{z}}}[\log(1-D(G(\mathbf{z})))],
\label{eq:gan_loss}
\end{equation}
where $\mathbf{x}$ is sampled from the data distribution $P_{data}$, $\mathbf{z}$ is sampled from a pre-defined noise distribution $P_{\mathbf{z}}$ (usually a standard normal or a standard uniform distribution), and $G(\mathbf{z})$ is the generator distribution. 
A trained generator thus can map from a predefined noise distribution to the distribution of designs. The noise input $\mathbf{z}$ is considered as the latent representation of the data, which can be used for design synthesis and exploration. Note that GANs often suffer from \textit{mode collapse}~\cite{salimans2016improved}, where the generator fails to capture all modes of the data distribution. In this work, by maximizing the diversity objective, mode collapse is discouraged as it leads to less diverse samples.

\subsection{Determinantal Point Processes}

Determinantal Point Processes (DPPs), which arise in quantum physics, are probabilistic models that model the likelihood of selecting a subset of diverse items as the determinant of a kernel matrix. Viewed as joint distributions over the binary variables corresponding to item selection, DPPs essentially capture negative correlations and provide a way
to elegantly model the trade-off between often competing notions of quality and diversity. The intuition behind DPPs is that the determinant of a kernel matrix roughly corresponds to the volume spanned by the vectors representing the items. Points that ``cover'' the space well should capture a larger volume of the overall space, and thus have a higher
probability. 

DPP kernels can be decomposed into quality and diversity parts~\cite{kulesza2012determinantal}. 
The $(i,j)$-th entry of a positive semi-definite DPP kernel $L$ can be expressed as:
\begin{equation}
 L_{ij} = q_i\;\phi(i)^T\;\phi(j)\;q_j.
 \label{eq:L_ij}
\end{equation}

We can think of $q_i \in R^+$ as a scalar value measuring the quality of an item $i$, and $\phi(i)^T\;\phi(j)$ as a signed measure of similarity between items $i$ and $j$. The decomposition enforces $L$ to be positive semidefinite. 
Suppose we select a subset $S$ of samples, then this decomposition allows us to write the probability of this subset $S$ as the square of the volume spanned by $q_i \phi _i$ for $i \in S$ using the equation below:
\begin{equation}\label{eq:eq4}
    \mathbb{P}_L(S)~\propto~\prod_{i \in S} ({q_i}^2) \det(K_S),
\end{equation}
where $K_S$ is the similarity matrix of $S$. 
As item $i$’s quality $q_i$ increases, so do the probabilities of sets containing item $i$. As two items $i$ and $j$ become more similar, ${\phi_i}^T \phi_j$ increases and the probabilities of sets containing both $i$ and $j$ decrease. The key intuition of MO-PaDGAN is that if we can integrate the probability of set selection from Eq.~(\ref{eq:eq4}) to the loss function of any generative model, then while training it will be encouraged to generate high probability subsets, which will be both diverse and high-performance.

\section{Problem Setting}

We consider a design represented by $\mathbf{x}\in\mathcal{X}$. In particular, $\mathbf{x}$ is a \textit{low-level} representation which does not encode any prior knowledge. For example, $\mathbf{x}$ can be a sequence of surface points or a pixel/voxel array when representing a geometric design. Since this low-level representation does not account for prior knowledge, it usually has a unnecessarily high dimension and any perturbation may lead to invalid designs with high probability. Thus, it is often impractical to directly control $\mathbf{x}$ in design optimization. Nonetheless, in many cases, we can find a \textit{parametric model} that parameterizes $\mathbf{x}$ by considering some prior knowledge. For example, we can use a B-spline curve to parameterize the surface of a geometry if it has a smooth surface. Using ``smoothness" as our prior knowledge, we constrain the space $\mathcal{X}$ so that some invalid designs cannot be represented (\ie, only a subset $\mathcal{X}'\subset\mathcal{X}$ can be represented). Assume that the parameters are $\mathbf{v}\in\mathcal{V}$ (\eg, $\mathbf{v}$ can be a sequence of B-spline control points) and the parametric model is a function $F: \mathcal{V}\rightarrow\mathcal{X}'$, now we can represent a design by a \textit{high-level} representation $\mathbf{v}$. We can easily obtain the low-level representation via $\mathbf{x} = F(\mathbf{v})$.

Given a parametric model $F$, a multi-objective design optimization problem can be formulated as:
\begin{equation*}
\begin{aligned}
\min_{\mathbf{v}} ~& f_m(F(\mathbf{v})), m=1,...,M \\
\text{s.t.} ~& g_j(F(\mathbf{v})) \leq 0, j=1,...,J \\
& h_k(F(\mathbf{v})) = 0, k=1,...,K \\
& v_d^{\text{inf}} \leq v_d \leq v_d^{\text{sup}}, d=1,...,D,
\label{eq:optimization_problem}
\end{aligned}
\end{equation*}
where $f$ denotes objective functions which compute performance indicators. Functions $g$ and $h$ compute inequality and equality constraints, respectively. $v_d^{\text{inf}}$ and $v_d^{\text{sup}}$ denote the lower and upper bounds, respectively, for the $d$-th parameter $v_d$.

The results of the optimization problem can be affected by two factors~\textemdash~the parametric model $F$ and the optimization algorithm. In this paper, we focus on the effects of the former. Given two parameterizations $F_1: \mathcal{V}_1\rightarrow\mathcal{X}'_1$ and $F_2: \mathcal{V}_2\rightarrow\mathcal{X}'_2$, $\mathcal{V}_1$ and $\mathcal{V}_2$ can have different dimensionality and $\mathcal{X}'_1$ and $\mathcal{X}'_2$ may cover different parts of $\mathcal{X}$. Such differences can contribute to different optimization results and are \textit{optimization algorithm agnostic}. To derive a parameterization that helps improve the performance of multi-objective design optimization, we propose to use a deep generative model, MO-PaDGAN, which can compactly represent designs with higher diversity and performances, as we will elaborate in the next section. We need the following two assumptions to build the MO-PaDGAN model: 
\begin{enumerate}
    \item There is a dataset containing possible design variants. This dataset is normally built by collecting historical designs, but can also be randomly generated following certain rules~\cite{chen2021deep}.
    \item One can evaluate the performance of any design (\eg, through simulations, experiments, visual inspection, etc.) or has a dataset containing design/performance pairs.
\end{enumerate}

In the next section, we explain the unique components in MO-PaDGAN and how to use it as a parameterization in design optimization.

\section{Methodology}
\label{sec:methodology}

MO-PaDGAN adds a \textit{performance augmented DPP loss} to a vanilla GAN architecture which measures the diversity and performance of a batch of generated designs during training. In this section, we first explain how to construct a DPP kernel. Then we show how to bake the DPP kernel into MO-PaDGAN's loss function. Finally, we formulate the multi-objective design optimization problem where we use MO-PaDGAN's generator as a parametric model and optimize designs over the learned latent space.

\subsection{Creating a DPP kernel}

We create the kernel $L$ for a sample of points generated by MO-PaDGAN from known inter-sample similarity values and performance vector. 

The similarity terms $\phi(i)^T \phi(j)$ can be derived using any similarity kernel, which we represent using $k(\mathbf{x}_i,\mathbf{x}_j) = \phi(i)^T \phi(j)$ and $\| \phi(i) \| = \| \phi(j) \| = 1$.
Here $\mathbf{x}_i$ is a vector representation of a design. Note that in a DPP model, the quality of an item is a scalar value representing design performance such as compliance, displacement, drag-coefficient.
For multivariate performance, we use a \textit{performance aggregator} to obtain a scalar quality
$q(\mathbf{x})=\mathbf{w}^T\mathbf{p}$, where $\mathbf{p}=(p_1,...,p_M)^T$ are $M$-dimensional performances and the corresponding weights $\mathbf{w}=(w_1,...,w_M)^T$ are positive numbers sampled uniformly at random and $\mathbf{1}^T\mathbf{w}=1$. The performance vector $\mathbf{p}$ can be estimated using a \textit{performance estimator} $E$, \ie, $\mathbf{p}=E(\mathbf{x})$. The performance estimator can be a physics simulator or a surrogate model.
Maximizing $q(\mathbf{x})$ pushes the non-dominated Pareto set of generated samples in the performance space to have higher values. While more complex performance aggregators (\eg, the Chebyshev distance from an ideal performance vector) are also applicable, we use the common linear scalarization to have fewer assumptions about the performance space. 

\subsection{Performance Augmented DPP Loss}

\begin{figure*}[!ht]
\centering
\includegraphics[width=1\textwidth]{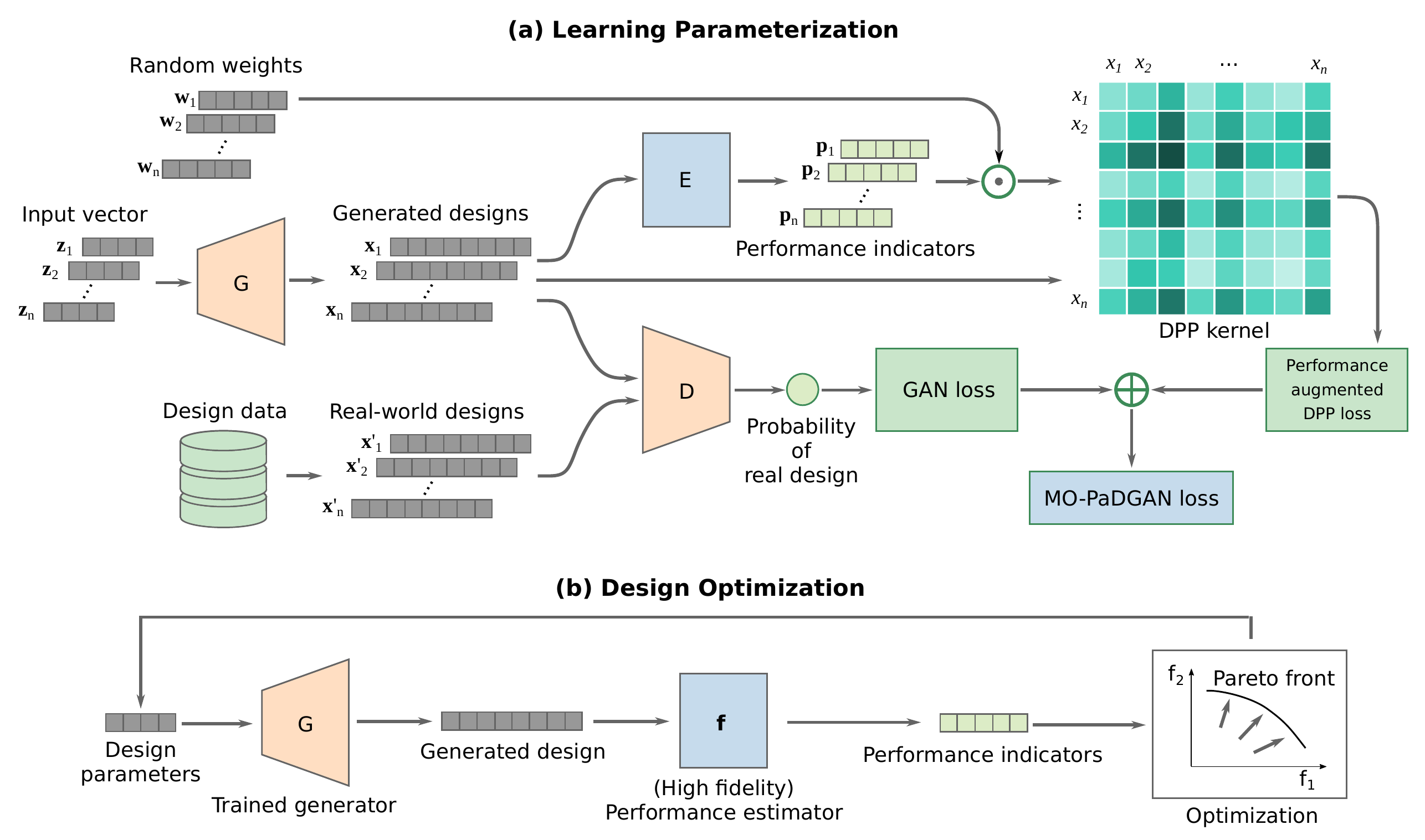}
\caption{Overview of our method: (a)~the MO-PaDGAN architecture at the training stage (The operator $\odot$ denotes performance aggregation); (b)~the multi-objective optimization loop at the design optimization stage.}
\label{fig:architecture}
\end{figure*}

The overall model architecture of MO-PaDGAN at the training stage is shown in Fig.~\ref{fig:architecture}(a). Our performance augmented DPP loss models diversity and performance simultaneously and gives a lower loss to sets of designs which are both high-performance and diverse. Specifically, we construct a kernel matrix $L_B$ for a generated batch $B$ based on Eq.~(\ref{eq:L_ij}). For each entry of $L_B$, we have
\begin{equation}
L_B(i,j) = k(\mathbf{x}_i,\mathbf{x}_j)\left(q(\mathbf{x}_i)q(\mathbf{x}_j)\right)^{\gamma_0},
\label{eq:L_B}
\end{equation}
where $\mathbf{x}_i,\mathbf{x}_j \in B$, $q(\mathbf{x})$ is the quality function at $\mathbf{x}$, and $k(\mathbf{x}_i,\mathbf{x}_j)$ is the similarity kernel between $\mathbf{x}_i$ and $\mathbf{x}_j$. 
To allow trade-off between quality and diversity, we adjust the dynamic range of the quality scores by using an exponent ($\gamma_0$) as a parameter to change the distribution of quality. A larger $\gamma_0$ increases the relative importance of quality as compared to diversity, which provides the flexibility to a user of MO-PaDGAN in deciding emphasis on quality vs diversity.

The performance augmented DPP loss is expressed as
\begin{equation}
\pazocal{L}_{\text{PaD}}(G) = -\frac{1}{|B|}\log\det(L_B) = -\frac{1}{|B|}\sum_{i=1}^{|B|} \log\lambda_i,
\label{eq:pad_loss}
\end{equation}
where $\lambda_i$ is the $i$-th eigenvalue of $L_B$. We add this loss to the vanilla GAN's objective in Eq.~(\ref{eq:gan_loss}) and form a new objective:
\begin{equation}
\min_G\max_D V(D,G) + \gamma_1 \pazocal{L}_{\text{PaD}}(G),
\label{eq:overall_loss}
\end{equation}
where $\gamma_1$ controls the weight of $\pazocal{L}_{\text{PaD}}$(G). 
For the backpropagation step, to update the weight $\theta_G^i$ in the generator in terms of $\pazocal{L}_{\text{PaD}}(G)$, we descend its gradient based on the chain rule:
\begin{equation}
\frac{\partial \pazocal{L}_{\text{PaD}}(G)}{\partial \theta_G^i} = \sum_{j=1}^{|B|} \left( \frac{\partial \pazocal{L}_{\text{PaD}}(G)}{\partial q(\mathbf{x}_j)} \frac{d q(\mathbf{x}_j)}{d \mathbf{x}_j} + \frac{\partial \pazocal{L}_{\text{PaD}}(G)}{\partial \mathbf{x}_j} \right) \frac{\partial \mathbf{x}_j}{\partial \theta_G^i},
\label{eq:gradient}
\end{equation}
where $\mathbf{x}_j = G(\mathbf{z}_j)$. Equation~(\ref{eq:gradient}) indicates a need for $dq(\mathbf{x})/d\mathbf{x}$, which is the gradient of the quality function. In practice, this gradient is accessible when the quality is evaluated through a performance estimator that is differentiable (\eg, using adjoint-based solver). If the gradient of a performance estimator is not available, one can either use numerical differentiation or approximate the quality function using a differentiable surrogate model (\eg, a neural network-based surrogate model, as used in our experiments).

\subsection{MO-PaDGAN Training and Evaluation}

The training of MO-PaDGAN follows the standard GAN training procedure, with the objective replaced by Eq.~(\ref{eq:overall_loss}). Specifically, in each iteration, we first update the discriminator $D$ by ascending its stochastic gradient, and then update the generator $G$ by descending its stochastic gradient. Please see Ref.~\cite{goodfellow2014generative} for details on the training of GANs.

Note that when using a data-driven surrogate model as the performance estimator $E$, it will perform unreliably on unrealistic designs, since it is usually trained on realistic data. This may cause problem at the beginning of MO-PaDGAN's training process, because the generator will produce unrealistic designs initially, which leads to unreliable estimator predictions and makes MO-PaDGAN training unstable. Ref.~\cite{chen2020padgan} proposed a few heuristics to address this problem. Please see Appendix~D for a detailed description.

There are standard metrics to evaluate a generative model in general (\eg, maximum mean discrepancy and Wasserstein distance between the data distribution and the generative distribution~\cite{xu2018empirical}). However, since MO-PaDGAN's goal is to build a better parameterization for design optimization, rather than approximating the data distribution, we can evaluate it by examining its design/performance space coverage and the downstream optimization task performance. We will show these results in Section.~\ref{sec:exp}.

\subsection{Optimizing Designs over Latent Representation}

The performance augmented DPP loss will encourage the generator to produce diverse and high-performance designs. After training, the generator $G$ forms a parametric model where the parameters are its input $\mathbf{z}$, as shown in Fig.~\ref{fig:architecture}(b). One can use $\mathbf{z}$ to synthesize new designs (\ie, $\mathbf{x}=G(\mathbf{z})$) which exhibit high diversity and performance. This indicates that the design space covered by MO-PaDGAN's generator is expanded and contains higher-performance designs, which is favorable when performing design optimization. The multi-objective optimization problem can be formulated as (when maximizing performances):
\begin{equation}
\begin{aligned}
\max_{\mathbf{z}} ~& f_m(G(\mathbf{z})), m=1,...,M \\
\text{s.t.} ~& g_j(G(\mathbf{z})) \leq 0, j=1,...,J \\
& h_k(G(\mathbf{z})) = 0, k=1,...,K \\
& z_d^{\text{inf}} \leq z_d \leq z_d^{\text{sup}}, d=1,...,D.
\label{eq:optimization}
\end{aligned}
\end{equation}
In this design optimization setting, variables $\mathbf{z}$ are treated as design parameters which controls the geometry of the design. The $d$-th design parameter $z_d$ is bounded by $z_d^{\text{inf}}$ and $z_d^{\text{sup}}$. As mentioned in Section~\ref{sec:gan}, since the distribution of $\mathbf{z}$ is defined as a prior, it is easy to set its lower and the upper bounds in design optimization. For example, if $\mathbf{z}$ has a standard uniform distribution, we can set the bounds of each design parameter to be [0,1]. The function $f_m$ is the estimator of the $m$-th performance indicator for a given design. In practice, it can be, for example, a Computational Fluid Dynamics (CFD) or Finite Element Method (FEM) solver, where we can evaluate the physics properties of any given design. Note that $\mathbf{f}=(f_1,...f_M)$ is not necessarily the same as the performance estimator $E$. To balance cost and accuracy, we can use a cheap low-fidelity model for $E$ and an expensive high-fidelity model for $\mathbf{f}$, since MO-PaDGAN training requires much more frequent performance evaluations whereas design optimization needs higher accuracy as it leads to the final product. Functions $g_j$ and $h_k$ define the $j$-th inequality and the $k$-th equality constraints, respectively. For example, $g_j$ can be the function that evaluates whether the bending stress in structural optimization exceeds the allowable stress, and $h_k$ can be the function that evaluates whether the lift in aerodynamic shape optimization equals the required lift.

There is no limitation of using any specific type of optimizer. Note that in the case where gradients are required (for gradient-based optimization), the functions $f_m$, $g_j$, and $h_k$ need to be differentiable (\eg, by using adjoint solvers). This allows access to the gradients of the objective and constraints with respect to the design $\mathbf{x}=G(\mathbf{z})$. Since we can already obtain $d\mathbf{x}/d\mathbf{z}$ using automatic differentiation, the gradients of the objective and constraints at $\mathbf{z}$ can be computed using the chain rule.

\section{Experimental Results}
\label{sec:exp}

In this section, we demonstrate the merit of using MO-PaDGAN as a parameterization in multi-objective optimization problems via two synthetic examples and a real-world airfoil design example. Please refer to our code and data for reproducing the experimental results\footnote{\url{https://github.com/wchen459/MO-PaDGAN-Optimization}}.

\subsection{Synthetic Examples}

\begin{figure}[t!]
\centering
\includegraphics[width=0.8\textwidth]{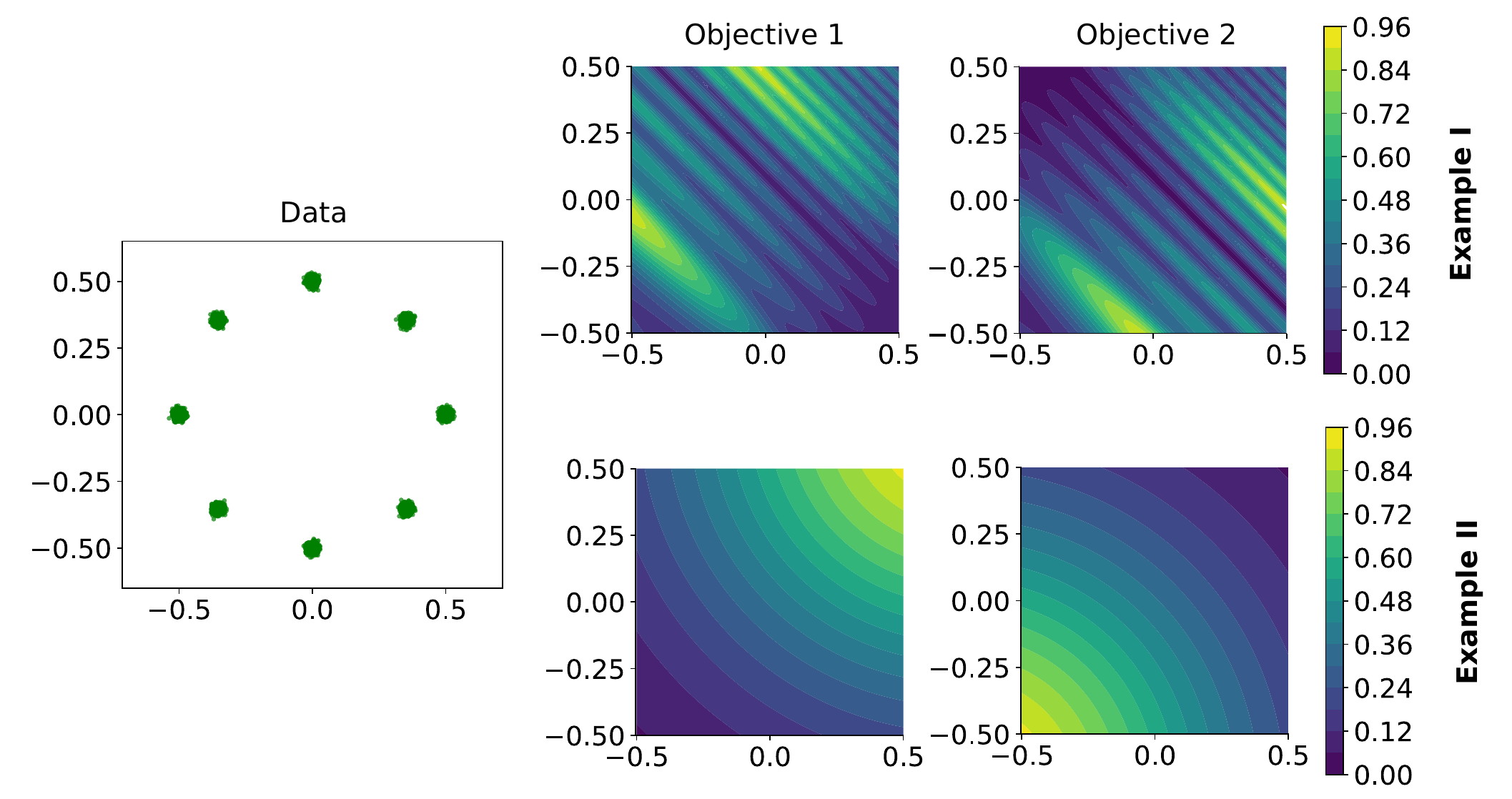}
\caption{Synthetic data and test quality (objective) functions.}
\label{fig:synthetic_data}
\end{figure}

\begin{figure*}[t!]
\centering
\includegraphics[width=1\textwidth]{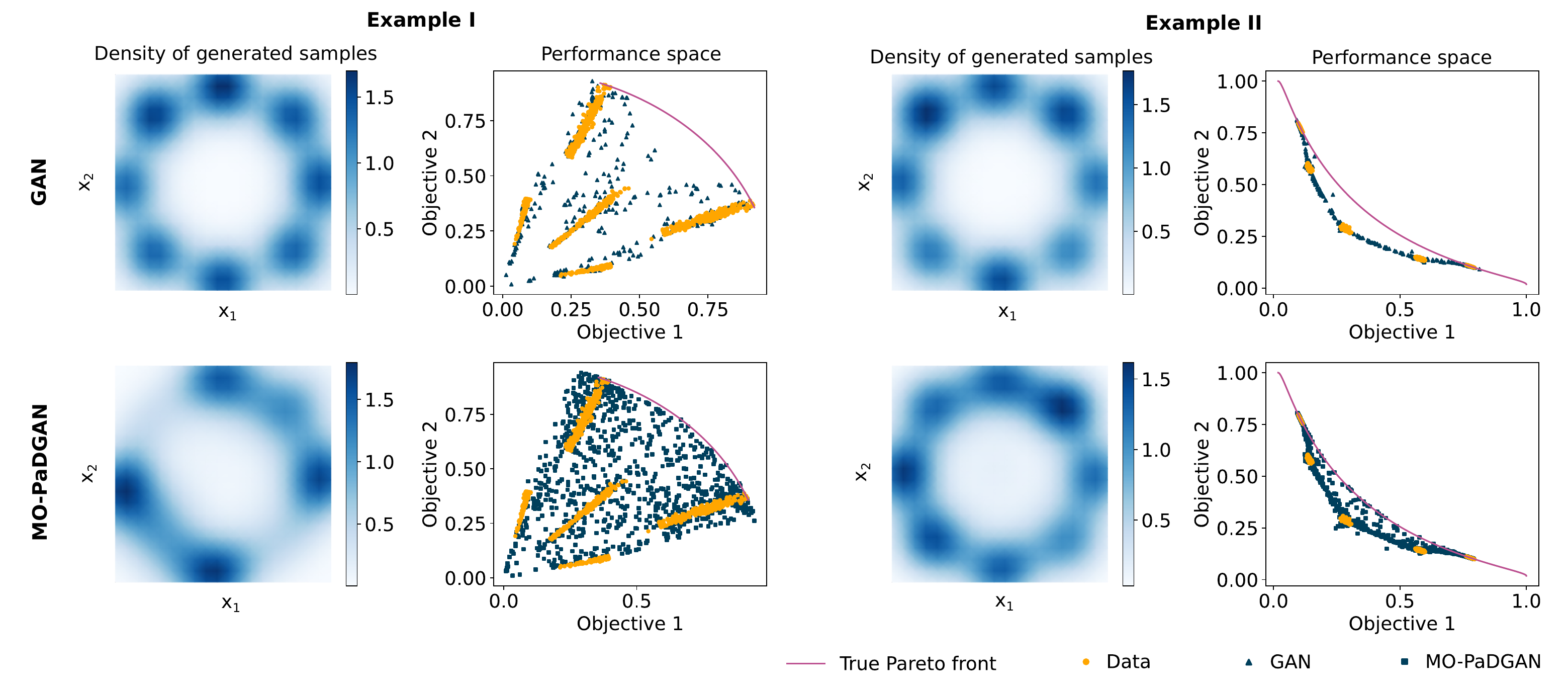}
\caption{Distribution and performance of generated points for the synthetic examples. While the data has few points near the true Pareto fronts, many of the generated samples from MO-PaDGAN are near the Pareto front.
}
\label{fig:synthetic_generated}
\end{figure*}

We create two synthetic examples, with eight clusters each, where cluster centers are evenly spaced around a circle (Fig.~\ref{fig:synthetic_data}). The sample size is 10,000 points. Both the synthetic examples have two parameters and two objectives, which enables us to visualize the results and compare the performance against a known ground truth Pareto optimal front. In the example following this section, we discuss a real-world airfoil example, which has larger dimensionality.

The objective functions of both synthetic examples are plotted in Fig.~\ref{fig:synthetic_data}. Please see Appendix~A for more details about the objective functions used.



We use the same network architecture and hyperparameter settings in both examples. We sample the noise vectors $\mathbf{z}$ from a two-dimensional normal distribution. The generator has four fully connected layers. Each hidden layer has 128 nodes and a LeakyReLU activation. We apply a hyperbolic tangent activation at the last layer. The discriminator has three fully connected layers. Each hidden layer has 128 nodes and a LeakyReLU activation. We apply a Sigmoid activation at the last layer. For MO-PaDGAN, we set $\gamma_0=2$ and $\gamma_1=0.5$. We use a RBF kernel with a bandwidth of 1 when constructing the DPP kernel matrix $L_B$ in Eq.~(\ref{eq:L_B}), \ie, $k(\mathbf{x}_i,\mathbf{x}_j)=\exp(-0.5\|\mathbf{x}_i-\mathbf{x}_j\|^2)$. 
We train both the MO-PaDGAN and the vanilla GAN for 10,000 iterations with 32 training samples drawn randomly from data at each iteration. For both generator and discriminator, we use Adam optimizer with a learning rate of $0.0001$.

\begin{figure}[t!]
\centering
\includegraphics[width=0.7\textwidth]{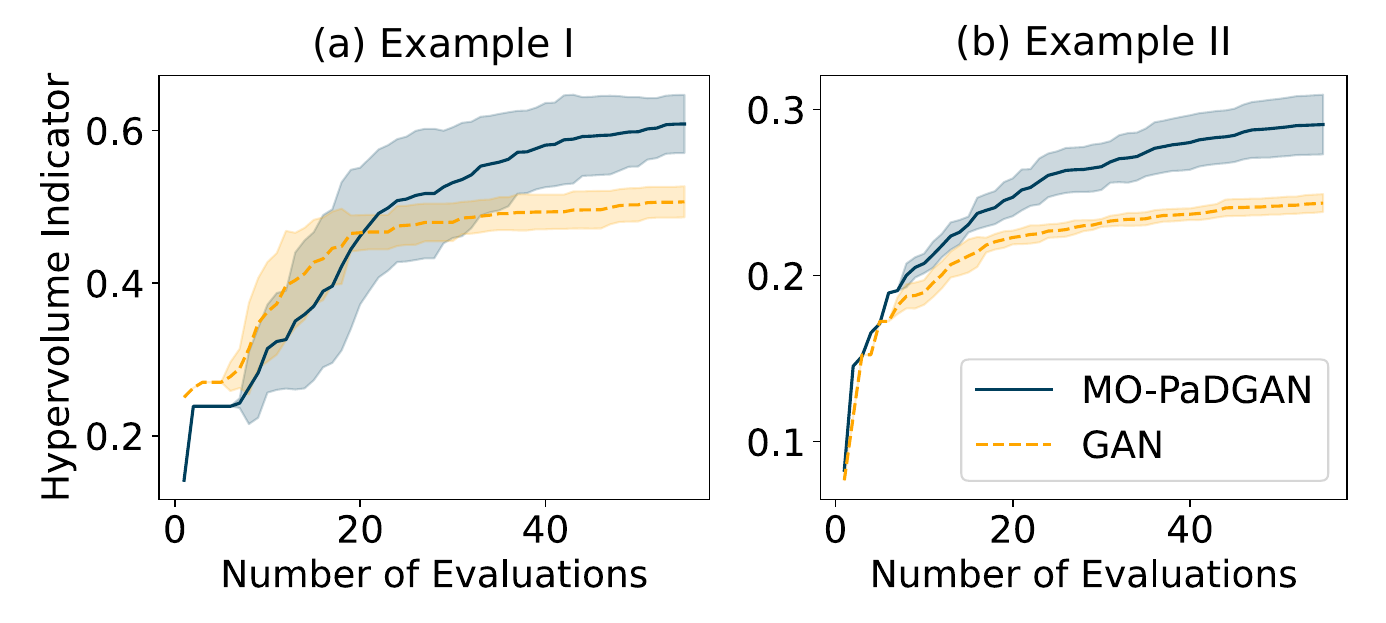}
\caption{Multi-objective optimization history for synthetic examples. Results shown are mean and standard deviation from ten optimization runs.}
\label{fig:synthetic_opt_history}
\end{figure}

\begin{figure}[t!]
\centering
\includegraphics[width=0.7\textwidth]{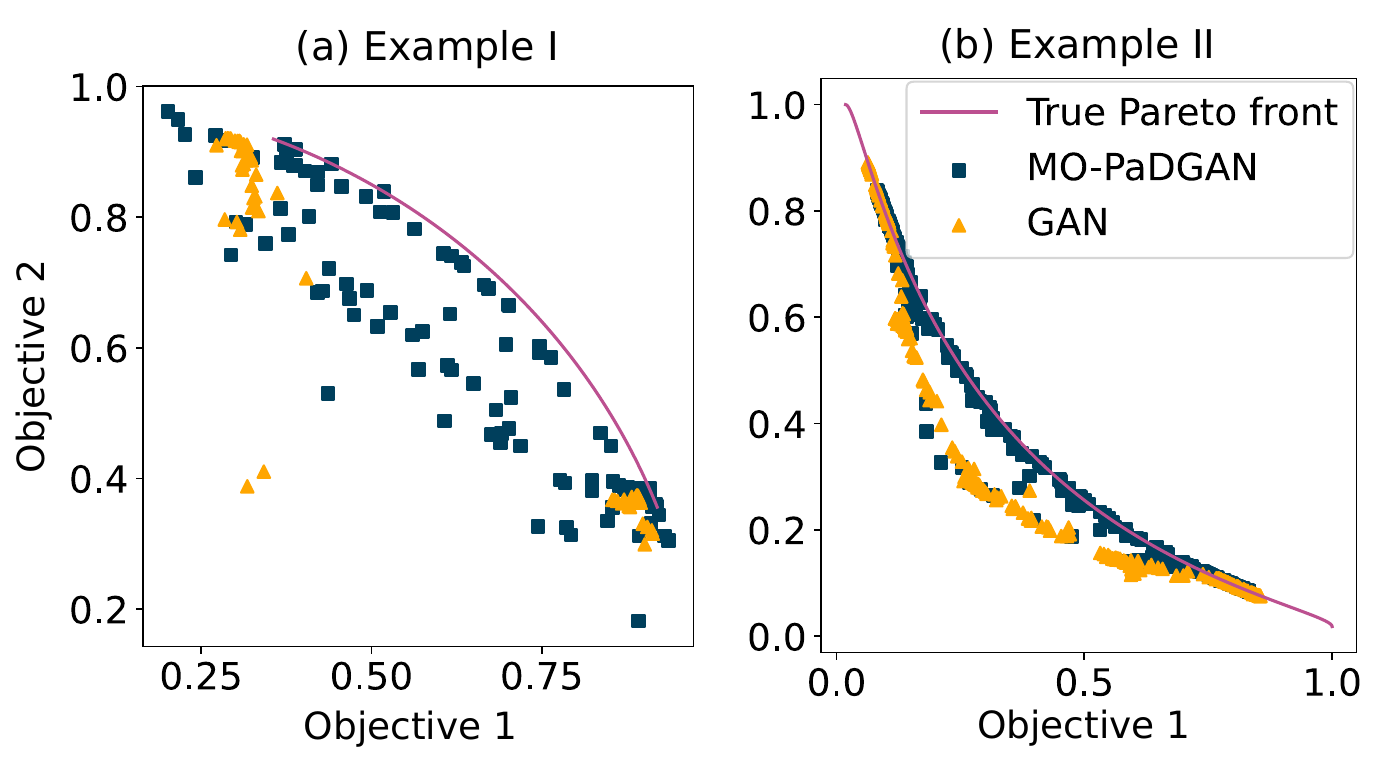}
\caption{The union of solution sets obtained by ten runs of multi-objective optimization for synthetic examples. MO-PaDGAN solutions are closer to the Pareto front than GAN.
}
\label{fig:synthetic_pareto_pts}
\end{figure}

Figure~\ref{fig:synthetic_generated} visualizes the parameter space and the performance (objective) space of points generated by a vanilla GAN and MO-PaDGAN, respectively. It shows that in the performance space, data are separated into several clusters, just as in the parameter space. The vanilla GAN approximates the data distribution well, except that it generates some points in between clusters, which is reasonable since the noise input $\mathbf{z}$ is continuous. Most points from the training data or generated by GAN are away from the (ground truth) Pareto fronts. In contrast, MO-PaDGAN generates points that fill up the gaps between clusters in the performance space and has many solutions near the Pareto fronts. This shows that guided by the performance augmented DPP loss, MO-PaDGAN pushes generated points towards higher performance in all objectives, although these new points are not seen in the dataset. Note that although results can differ due to the stochasticity introduced in the neural network training process (\eg, random weights initialization and stochastic gradients), we do not observe any qualitative difference on the results.

We then perform multi-objective Bayesian optimization (MOBO)~\cite{couckuyt2014fast} to solve Eq.~(\ref{eq:optimization}) while using the trained generators as parameterizations. Here the variables are the generator's two-dimensional input vector $\mathbf{z}$. The objective functions $f_1$ and $f_2$ are the ones shown in Fig.~\ref{fig:synthetic_data} (please see Appendix~A for details). Note that here we use synthetic functions as the objectives only for demonstration and validation purpose. In reality, the objectives can be any quantities of interest (e.g., displacement, lift coefficient, weight, cost, etc.). The bounds $z_d^{\text{inf}}$ and $z_d^{\text{sup}}$ ($d=1,2$) are -2 and 2, which are two times the standard deviation of the distribution of $z_d$. In these two examples, we do not set constraints $g$ and $h$. We sample 5 points (noise input vectors) via Latin Hypercube Sampling (LHS)~\cite{mckay2000comparison} for initial evaluations. This is followed by Bayesian optimization to select 50 additional evaluations. Bayesian optimization uses a Gaussian process to model the objective function and its uncertainty~\cite{rasmussen2006gaussian}. We use a Mat\'ern 5/2 kernel for the Gaussian process. Please see Ref.~\cite{rasmussen2006gaussian} and our open-sourced code for more implementation details. We use the hypervolume indicator~\cite{zitzler1999multiobjective} to assess the quality of a solution set. The hypervolume measures the volume of the dominated portion of the performance space bounded from below by a reference point. We use $(0,0)$ as the reference points for both examples. Figure~\ref{fig:synthetic_opt_history} shows the history of hypervolume indicators during optimization from ten runs. Figure~\ref{fig:synthetic_pareto_pts} shows the union of final solution sets from these ten runs. The results indicate that MO-PaDGAN achieves a much better performance than the vanilla GAN in discovering the Pareto front. This is expected since, as shown in Fig.~\ref{fig:synthetic_generated}, the parameterization learned by MO-PaDGAN can represent high-performance samples near the true Pareto front, whereas a vanilla GAN cannot.

\subsection{Airfoil Design Example}

\begin{figure*}[t!]
\centering
\includegraphics[width=1\textwidth]{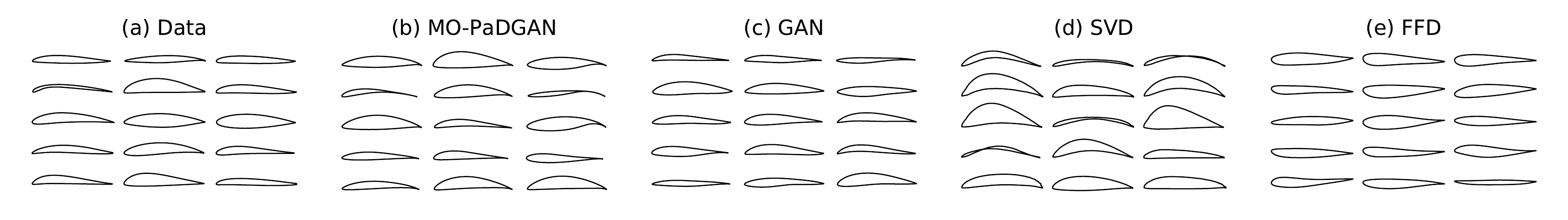}
\caption{Airfoil shapes randomly drawn from the training data or generated by parameterizations.}
\label{fig:airfoil_samples}
\end{figure*}


An airfoil is the cross-sectional shape of an airplane wing or a propeller/rotor/turbine blade. Airfoil shape optimization is crucial when designing a wing or a blade, as we usually optimize multiple 2D airfoils at different cross-sections and interpolate between sections to get the full 3D geometry. In this example, we use the UIUC airfoil database\footnote{\url{http://m-selig.ae.illinois.edu/ads/coord_database.html}} as our data source. It provides the geometries of nearly 1,600 real-world airfoil designs. We preprocessed and augmented the dataset similar to \cite{chen2020airfoil}, which led to a dataset of 38,802 airfoils, each of which is represented by 192 surface points (\ie, $\mathbf{x}_i\in \mathbb{R}^{192\times 2}$). Figure~\ref{fig:airfoil_samples}(a) shows airfoil shapes randomly drawn from the training data. We use two performance indicators for designing the airfoils~\textemdash~the lift coefficient ($C_L$) and the lift-to-drag ratio ($C_L/C_D$). These two are common objectives in aerodynamic design optimization problems and have been used in different multi-objective optimization studies~\cite{park2010optimal}. We use XFOIL~\cite{drela1989xfoil} for CFD simulations and compute $C_L$ and $C_D$ values\footnote{We set $C_L=C_L/C_D=0$ for unsuccessful simulations.}. We scaled the performance scores between 0 and 1. To provide the gradient of the quality function for Eq.~(\ref{eq:gradient}), we trained a neural network-based surrogate model on all 38,802 airfoils to approximate both $C_L$ and $C_D$.

We use a residual neural network (ResNet)~\cite{he2016deep} as the surrogate model to predict the performance indicators and a B\'ezierGAN~\cite{chen2020airfoil,chen2019aerodynamic,chen2018bezier} to parameterize airfoils. Please refer to Appendix~B and the code for details on their network architectures. Different from the vanilla GAN's architecture, B\'ezierGAN's generator has two inputs~\textemdash~the \textit{latent code} $\mathbf{z}$ and the noise vector $\mathbf{z}'$, which follows the InfoGAN's setting~\cite{chen2016infogan}. The additional latent code provides a more disentangled representation and hence will be used as our design parameters. We set the latent dimension and the noise dimension to 5 and 10, respectively. We sample the latent codes from a uniform distribution, \ie, $z_i\sim U(0,1)$, $i=1,...,5$. We sample the noise vectors from a normal distribution, \ie, $z_j'\sim \mathcal{N}(0,1)$, $j=1,...,10$. For simplicity, we refer to the B\'ezierGAN as a vanilla GAN and the B\'ezierGAN with the loss $\pazocal{L}_{PaD}$ as a MO-PaDGAN in the rest of the paper. 
We follow the same training configuration (\ie, training iterations, batch size, optimizer, and learning rate) as in the synthetic examples. For MO-PaDGAN, we set $\gamma_0=5$ and $\gamma_1=0.2$. We use the RBF kernel with a bandwidth of 1 in the DPP kernel matrix. 



\begin{figure}[t!]
\centering
\includegraphics[width=0.7\textwidth]{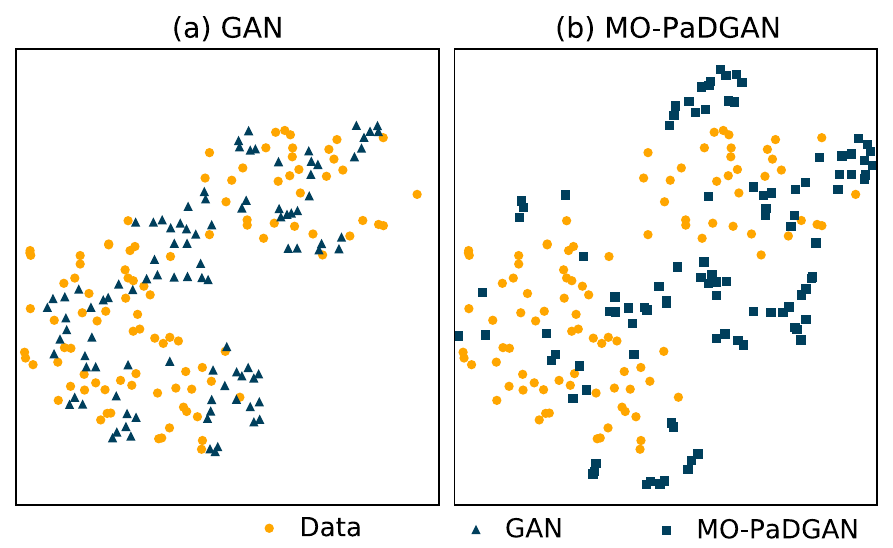}
\caption{Randomly sampled airfoils embedded into a 2D space via t-SNE.} 
\label{fig:airfoils_tsne}
\end{figure}

\begin{figure}[t!]
\centering
\includegraphics[width=0.7\textwidth]{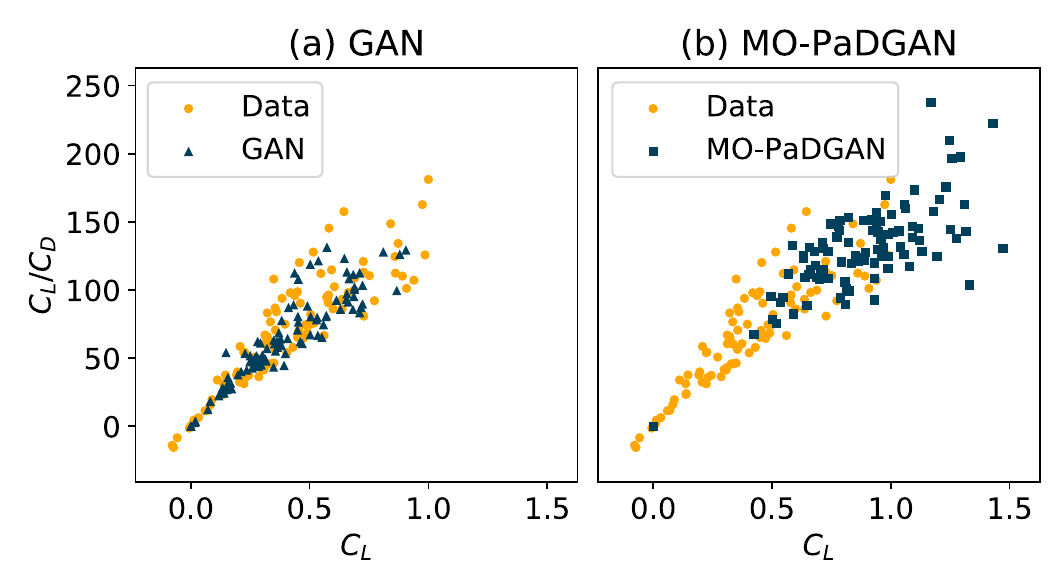}
\caption{Performance space visualization for airfoil samples shown in Fig.~\ref{fig:airfoils_tsne}.}
\label{fig:airfoils_perf}
\end{figure}


Airfoils randomly generated by MO-PaDGAN and GAN are shown in Fig.~\ref{fig:airfoil_samples}(b) and (c), respectively. Both MO-PaDGAN and GAN can generate realistic airfoil shapes. To compare the distribution of real and generated airfoils in the design space, we map randomly sampled airfoils into a two-dimensional space through t-SNE, as shown in Figure~\ref{fig:airfoils_tsne}. The results indicate that compared to a vanilla GAN, MO-PaDGAN can generate airfoils that are outside the boundary of the training data, driven by the tendency to maximize diversity using the DPP loss. Figure~\ref{fig:airfoils_perf} visualizes the joint distribution of $C_L$ and $C_L/C_D$ for randomly sampled/generated airfoils. It shows that MO-PaDGAN generates airfoils with performances exceed randomly sampled airfoils from training data and the vanilla GAN (\ie, the non-dominated Pareto set of generated samples is pushed further in the performance space to have higher values). Hence, Figures~\ref{fig:airfoils_tsne} and \ref{fig:airfoils_perf} indicate that MO-PaDGAN can expand the existing boundary of the design space towards high-performance regions outside the training data. This directed expansion is allowed since the MO-PaDGAN's generator is updated with the quality gradients (\ie, $dq(\mathbf{x})/d\mathbf{x}$ in Eq.~(\ref{eq:gradient})). 

We then perform multi-objective Bayesian optimization (MOBO) and multi-objective evolutionary algorithm (MOEA, C-TAEA in particular)~\cite{li2018two} to solve Eq.~(\ref{eq:optimization}) while using the trained generator $G$ as the parameterization. We use only the latent codes as design parameters (variables) and fix the noise vector to zeros. The objectives are $C_L$ and $C_L/C_D$. Thus, $f_1$ and $f_2$ in Eq.~(\ref{eq:optimization}) both contain the CFD simulator. The bounds $z_d^{\text{inf}}$ and $z_d^{\text{sup}}$ ($d=1,...,5$) are 0 and 1, which are the bounds of the uniform distribution of $z_d$. In this example, we do not set constraints $g$ and $h$. For MOBO, we sample 15 points (latent codes) via LHS for initial evaluations and 150 subsequent evaluations for Bayesian optimization. We use the same settings for Gaussian process as in the synthetic examples. To assess the optimization performance, we use $(0,0)$ as the reference points in the hypervolume indicator. For MOEA, we have 11 generations with a population size of 15. Therefore, MOEA has the same number of evaluations as MOBO. The number of offsprings is 1. We use the simulated binary crossover~\cite{deb2007self} with a distribution index of 30 and the polynomial mutation~\cite{deb1996combined} with a distribution index of 20. Please refer to our code for detailed implementation.

We also compare MO-PaDGAN and GAN with other two state-of-the-art airfoil parameterizations~\textemdash~Singular Value Decomposition (SVD)~\cite{poole2015metric,poole2019efficient} and Free-Form Deformation (FFD)~\cite{masters2017geometric}. The SVD extracts salient airfoil deformation modes from a dataset using truncated SVD and uses the weights of those modes as design parameters. We set the number of extracted modes to 5 in this experiment (\ie, 5 design parameters). The FFD represents a new shape by deforming some initial shape via moving a set of $m\times n$ control points. We set the number of control points to $3\times 4$ and only move their $y$ coordinates (\ie, 12 design parameters). We refer interested readers to our code for detailed algorithms and settings of these two parameterizations. Figure~\ref{fig:airfoil_samples}(d) and (e) shows randomly synthesized airfoils using SVD and FFD. Compared to airfoils drawn from data or generated by MO-PaDGAN/GAN, airfoils synthesized by SVD/FFD either have unrealistic (\eg, self-intersecting) shapes or insufficient diversity. This is expected since, as generative models, the objective of GAN and its variants is to learn data distribution, so it is able to generate designs with similar (realistic) appearance and diversity to the data if trained properly, whereas non-generative models do not capture the feasible boundary.

\begin{figure}[t!]
\centering
\includegraphics[width=0.8\textwidth]{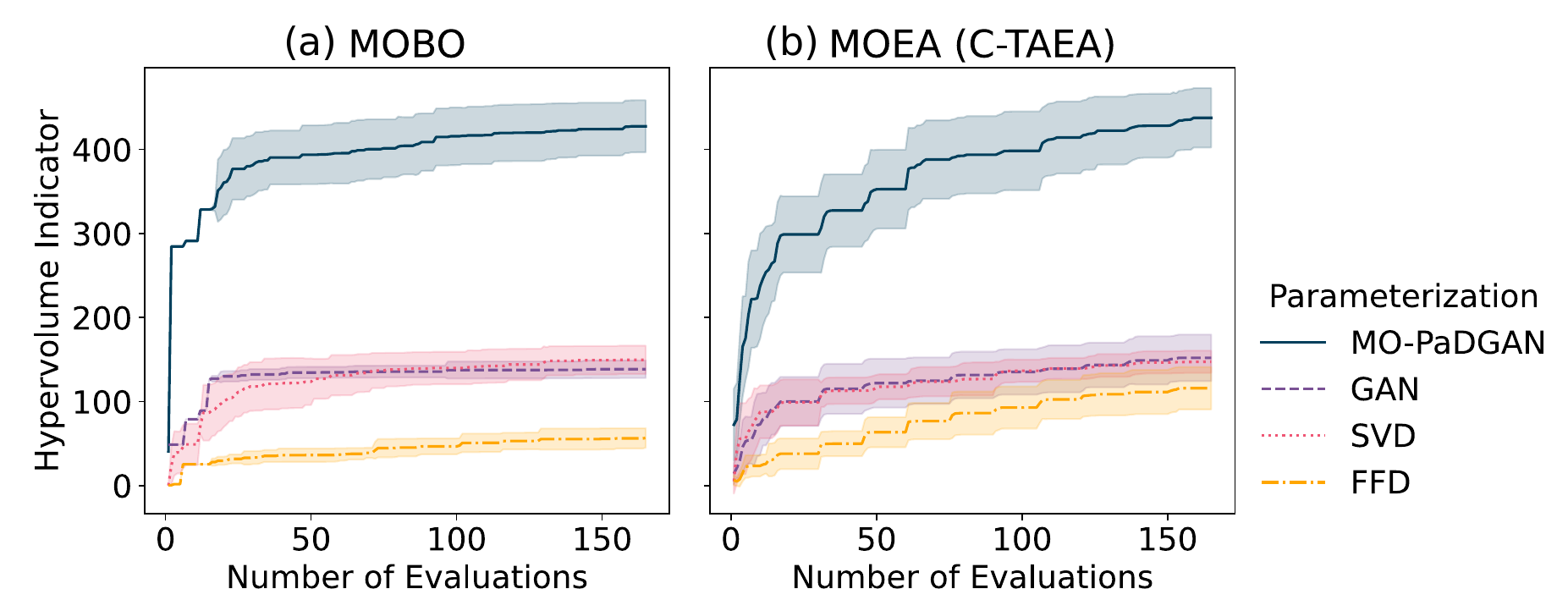}
\caption{Hypervolume indicator history for airfoil optimization using different parameterizations. Results shown are mean and standard deviation from ten optimization runs.
MO-PaDGAN achieves on average an over 180\% improvement in the final hypervolume indicator compared to the second-best parameterization.
}
\label{fig:airoil_opt_history}
\end{figure}

\begin{figure}[t!]
\centering
\includegraphics[width=0.8\textwidth]{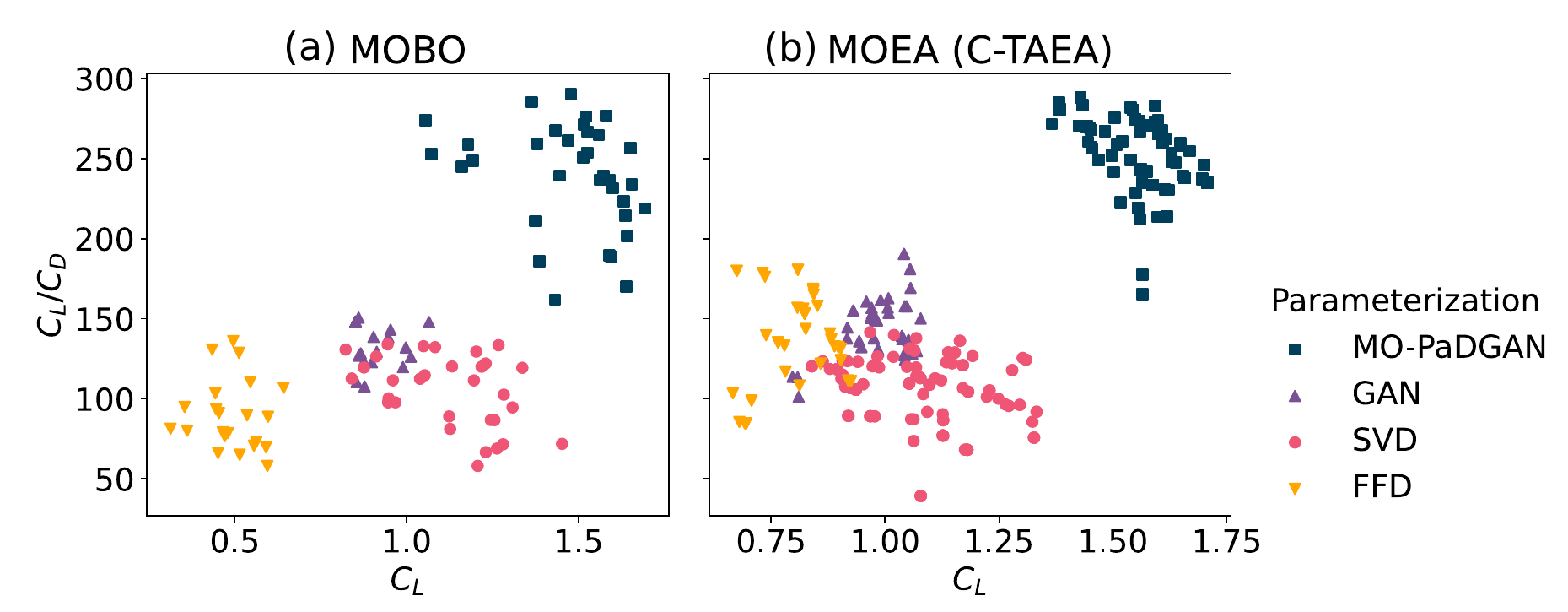}
\caption{The union of solution sets obtained by ten runs of multi-objective airfoil design optimization using MO-PaDGAN, GAN, SVD, and FFD. We see that MO-PaDGAN solutions dominate solutions from all other methods.}
\label{fig:airfoil_pareto_pts}
\end{figure}

\begin{figure}[t!]
\centering
\includegraphics[width=0.75\textwidth]{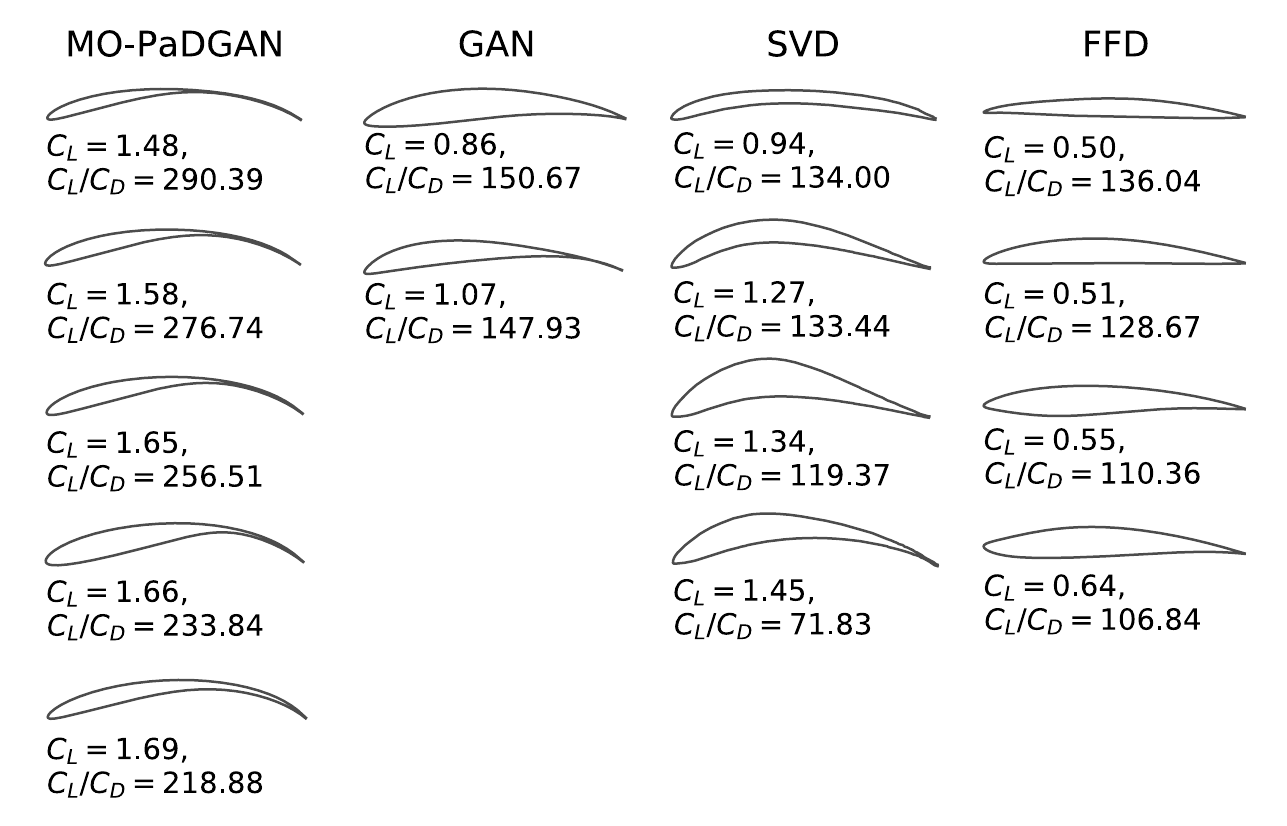}
\caption{Non-dominated set among MOBO solutions shown in Fig.~\ref{fig:airfoil_pareto_pts}(a). We see that MO-PaDGAN outperforms other parameterizations on both objectives.}
\label{fig:airfoil_pareto_airfoils_bo}
\end{figure}

\begin{figure}[p]
\centering
\includegraphics[width=0.75\textwidth]{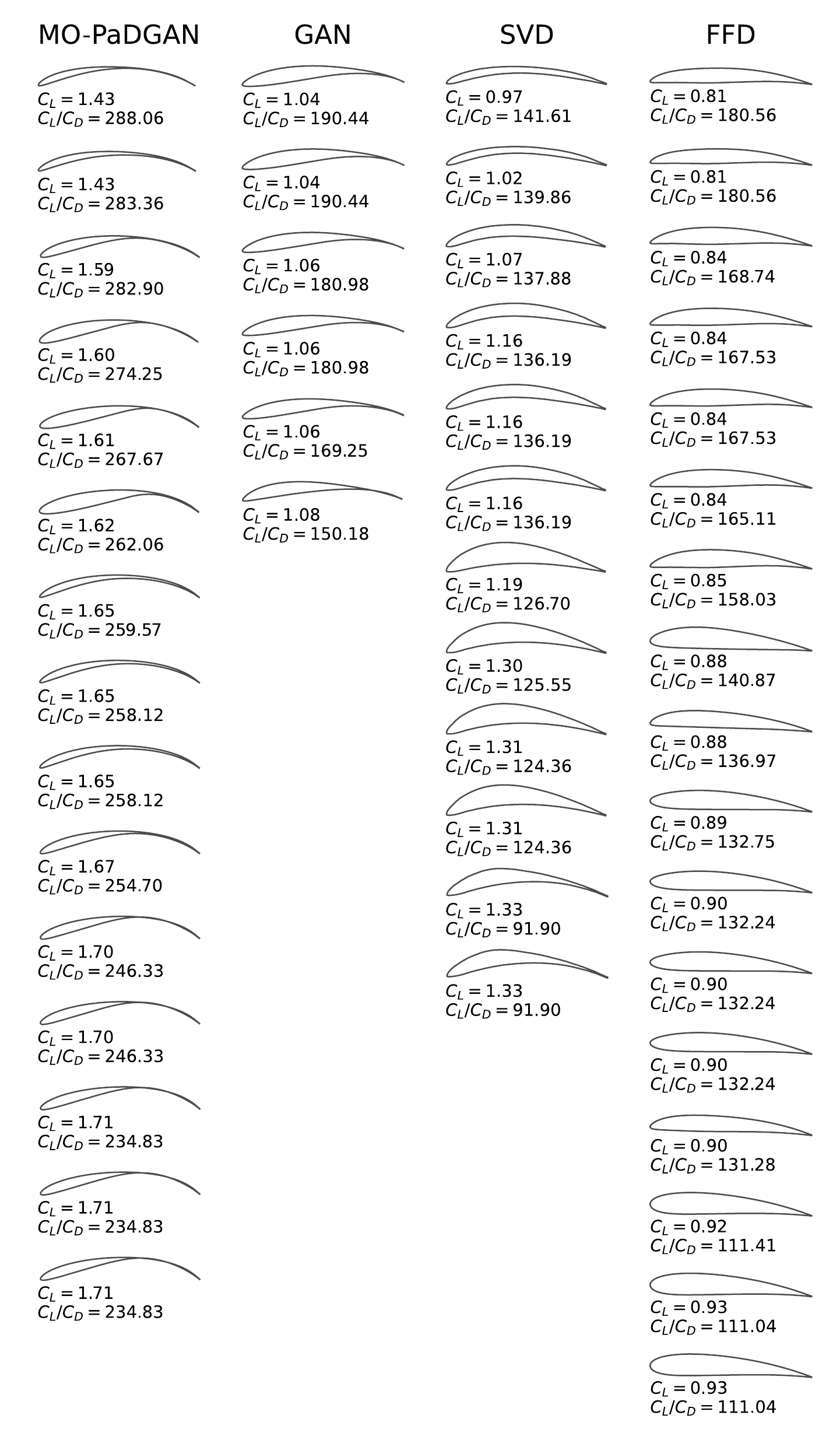}
\vspace*{-6mm}
\caption{Non-dominated set among MOEA solutions shown in Fig.~\ref{fig:airfoil_pareto_pts}(b). We see that MO-PaDGAN outperforms other parameterizations on both objectives (numbers are rounded to the nearest hundreds).}
\label{fig:airfoil_pareto_airfoils_ea}
\end{figure}

\begin{figure}[t!]
\centering
\includegraphics[width=0.8\textwidth]{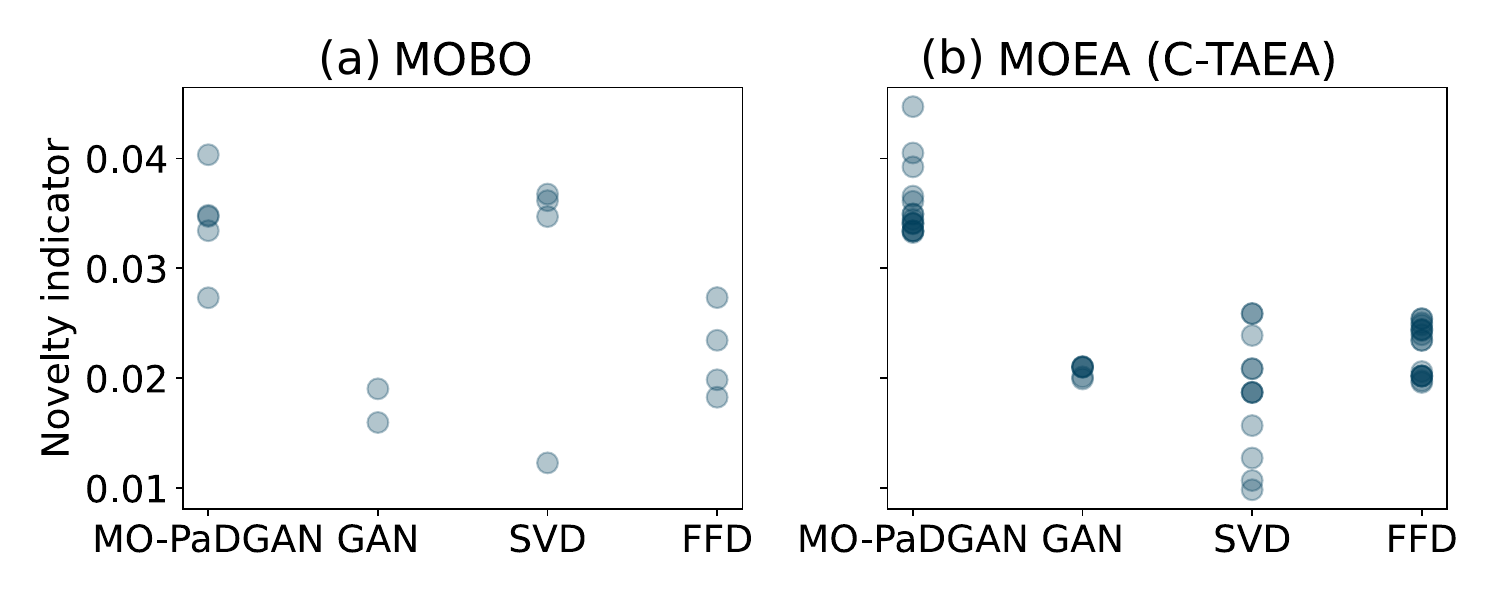}
\caption{Novelty indicators of airfoils shown in Fig.~\ref{fig:airfoil_pareto_airfoils_bo} and \ref{fig:airfoil_pareto_airfoils_ea}. This indicates that MO-PaDGAN solutions are most novel in general. Note that SVD and FFD are also able to give relatively novel solutions, but cannot reach the same level of performance as MO-PaDGAN based on Fig.~\ref{fig:airfoil_pareto_airfoils_bo} and \ref{fig:airfoil_pareto_airfoils_ea}.}
\label{fig:airfoil_pareto_novelty}
\end{figure}

Figure~\ref{fig:airoil_opt_history} shows that, when using a MO-PaDGAN for parameterization, the hypervolume indicators are significantly higher than other parameterizations throughout the whole optimization process. This observation is consistent using either MOBO or MOEA as the optimizer. This is expected since the main contributor to MO-PaDGAN's exceptional optimization results is its ability as a parameterization to expand the design space towards high-performance regions, and this ability is independent of any downstream optimization solvers (\ie, optimization algorithm agnostic). The MO-PaDGAN achieves on average an over 180\% improvement in the final hypervolume indicator compared to the second-best parameterization. Figure~\ref{fig:airfoil_pareto_pts} shows the union of solution sets obtained by ten optimization runs, which illustrates a notable gain in the performances of optimized MO-PaDGAN airfoils, when optimizing using either MOBO or MOEA. We further obtain the non-dominated Pareto set for each parameterization given the points shown in Fig.~\ref{fig:airfoil_pareto_pts} and plot the shapes of airfoils in the Pareto sets in Figs.~\ref{fig:airfoil_pareto_airfoils_bo} and \ref{fig:airfoil_pareto_airfoils_ea}. It shows that solutions from MO-PaDGAN's Pareto set dominate all the solutions in other Pareto sets. To assess whether the optimized shapes are novel, \ie, different from any existing shapes (from the dataset), we compute the \textit{novelty indicator} for each optimized shape from Figs.~\ref{fig:airfoil_pareto_airfoils_bo} and \ref{fig:airfoil_pareto_airfoils_ea}. Specifically, the novelty indicator is represented by the difference between each optimized shape and the most similar shape (\ie, the nearest neighbor) from the original dataset. While the difference can be measured with various metrics based on the specific application, in this experiment we use Hausdorff distance, which measures the distance between any two sets of airfoil surface points. The novelty indicators are shown in Fig.~\ref{fig:airfoil_pareto_novelty}. It shows that while GAN, as a generative model, can generate realistic designs which exhibit similar diversity as data, it is harder to discover novel solutions in design optimization, compared to SVD or FFD. In contract, also as a generative model, MO-PaDGAN can discover optimal solutions with much higher novelty. This shows that the solutions found on the trade-off front are far from the original dataset. 
Because by simultaneously modeling diversity and quality, MO-PaDGAN encourages high-performance designs to be diverse and hence expands the space of high-performance designs.

\section{Conclusion}

We proposed MO-PaDGAN to encourage diverse and high-quality sample generation, where the quality of a sample can be specified by a multivariate metric. This is particularly useful for Engineering Design applications, which often require simultaneous improvement in several performance indicators. By using the new model as a design parameterization, we showed that it expands the design space towards high-performance regions, whereas a vanilla GAN only generates designs within the original design space bounded by data. As a result, MO-PaDGAN can represent designs that have higher-performance but also are novel compared to existing ones. We further used MO-PaDGAN as a new parameterization in multi-objective optimization tasks. Through two synthetic examples, we found that MO-PaDGAN allows the discovery of the underlying full Pareto fronts even though data do not cover those Pareto fronts. In the airfoil design example, we compared MO-PaDGAN with three state-of-the-art parameterization methods~\textemdash~a vanilla GAN, SVD, and FFD. We showed that MO-PaDGAN achieves on average an over 180\% improvement in the hypervolume indicator compared to the second-best parameterization. We also discovered that the Pareto-optimal solutions by using MO-PaDGAN exhibit higher novelty than GAN.

Although we demonstrated the real-world application of MO-PaDGAN through an airfoil design example, this model shows promise in extending to other Engineering problems like material microstructural design and molecule discovery. The proposed model also generalizes to other generative models like VAEs, where we can add the performance augmented DPP loss to the Kullback-Leibler divergence and the reconstruction loss. 

Future work will investigate questions focused on two key limitations of the method~\textemdash~a)~How to extend the method to domains where gradient information of design performance is not available, and b)~How to capture the entire Pareto front uniformly for many-objective optimization problems.

\clearpage
\begin{center}
\textbf{\Large Appendix A. Synthetic Examples: Objective Functions}
\end{center}
\setcounter{equation}{0}
\setcounter{figure}{0}
\setcounter{table}{0}
\setcounter{section}{0}
\makeatletter
\renewcommand{\theequation}{A.\arabic{equation}}
\renewcommand{\thefigure}{A.\arabic{figure}}
\renewcommand{\thetable}{A.\arabic{table}}
\renewcommand{\thesection}{A.\arabic{section}}

In Example~\RNum{1}, we use the modified KNO1~\cite{knowles2006parego} as the objective functions:
\begin{align*}
\max_{x_1', x_2'} ~& f_1 = (r/20)\cos(\phi) \\
\max_{x_1', x_2'} ~& f_2 = (r/20)\sin(\phi) \\
\text{s.t.} ~& r = 9 - \left[3\sin\left(\frac{5}{2(x_1+x_2)^2}\right) + 3\sin(4(x_1+x_2))\right. \\
& ~~~~~~~\left. + 5\sin(2(x_1+x_2)+2)\vphantom{\frac{5}{2(x_1+x_2)^2}}\right] \\
& \phi = \frac{\pi(x_1-x_2+3)}{12} \\
& x_1 = 3(x_1'+0.5) \\
& ~x_2 = 3(x_2'+0.5) \\
& x_1', x_2' \in [-0.5, 0.5].
\end{align*}
The Pareto front for this function lies on the line defined by $x_1+x_2=0.4705$ (in the parametric space). There are multiple locally optimal fronts.

In Example~\RNum{2}, we use the modified VLMOP2~\cite{van1999multiobjective} as the objective functions:
\begin{align*}
\max_{x_1, x_2} ~& f_1 = \exp\left(-\left(x_1-\frac{1}{\sqrt{2}}\right)^2-\left(x_2-\frac{1}{\sqrt{2}}\right)^2\right) \\
\max_{x_1, x_2} ~& f_2 = \exp\left(-\left(x_1+\frac{1}{\sqrt{2}}\right)^2-\left(x_2+\frac{1}{\sqrt{2}}\right)^2\right) \\
\text{s.t.} ~& x_1, x_2 \in [-0.5, 0.5].
\end{align*}
The Pareto front for this bi-objective problem is concave (in the objective space) and lies on the line defined by $x_1-x_2=0$ (in the parametric space).

For simplicity, both performance estimators $E$ (used in the training stage) and $\mathbf{f}$ (used in the multi-objective optimization stage) are defined by the above objective functions. In practice, one can use a lower-fidelity but differentiable $E$ during MO-PaDGAN training and a higher-fidelity $\mathbf{f}$ during design optimization, as shown in the airfoil design example.

\clearpage
\begin{center}
\textbf{\Large Appendix B. Airfoil Design Example: Neural Network Surrogate Model}
\end{center}
\setcounter{equation}{0}
\setcounter{figure}{0}
\setcounter{table}{0}
\setcounter{section}{0}
\makeatletter
\renewcommand{\theequation}{B.\arabic{equation}}
\renewcommand{\thefigure}{B.\arabic{figure}}
\renewcommand{\thetable}{B.\arabic{table}}
\renewcommand{\thesection}{B.\arabic{section}}

\begin{figure}[t!]
\centering
\includegraphics[width=0.7\textwidth]{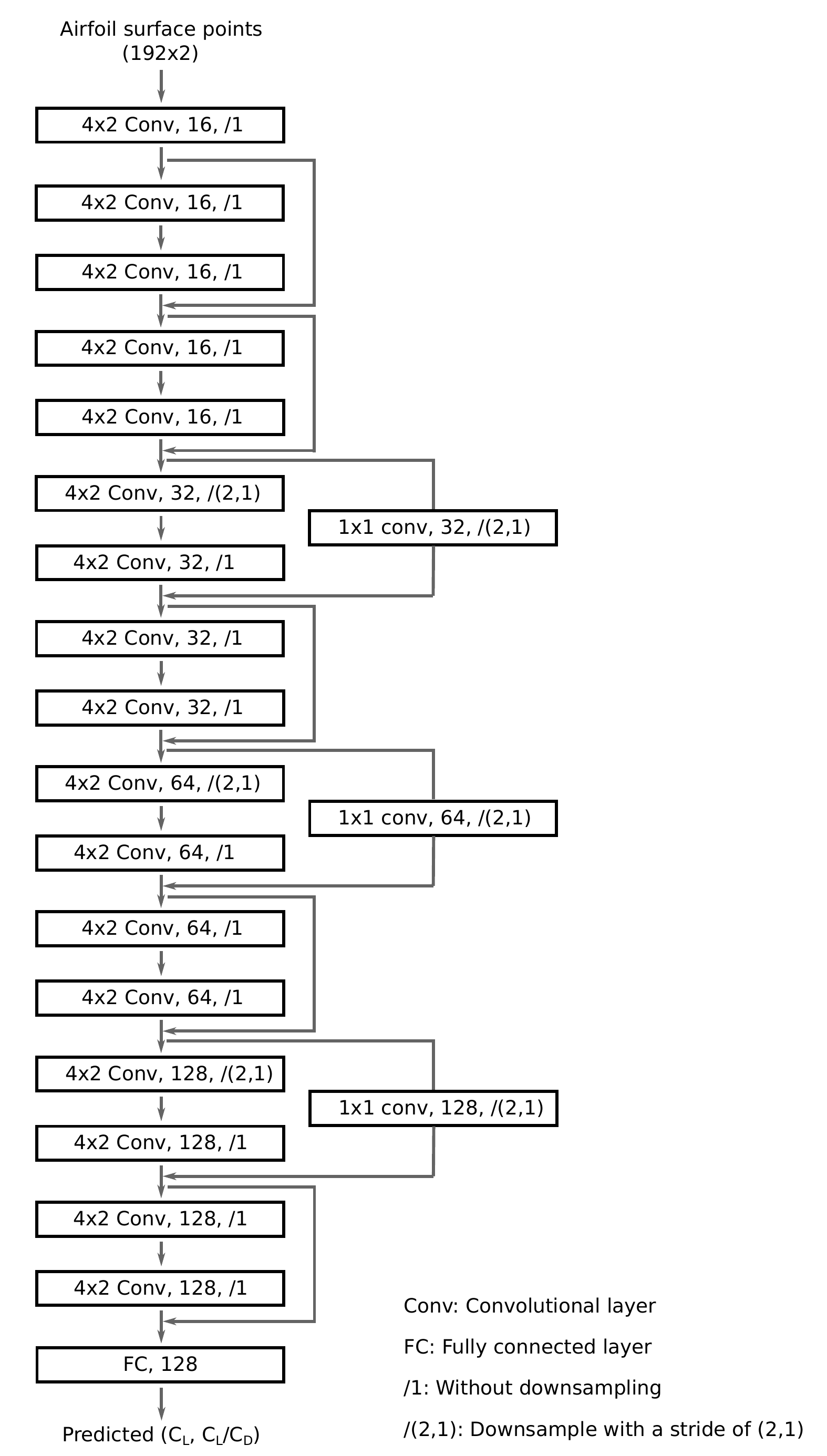}
\caption{Architecture of the neural network surrogate model for airfoil performance prediction.}
\label{fig:app_surrogate}
\end{figure}

To obtain a differentiable performance estimator $E$ for airfoil designs, we train a ResNet as a surrogate model of a higher-fidelity physics solver. The model architecture is shown in Fig.~\ref{fig:app_surrogate}. We use eight residual blocks, three of which has down-sampling. The convolution is performed along the sequence of airfoil surface point coordinates. We use batch normalization and a leaky ReLU activation following each hidden layer, and a Sigmoid activation at the output layer. We use XFOIL to compute the target $C_L$ and $C_L/C_D$ values of the airfoil data and scale the target values between 0 and 1.

We use 38,802 airfoil designs as the training data. We train the surrogate model for 10,000 iterations with a batch size of 256 at each iteration. We use the Adam optimizer with a learning rate of 0.0001, $\beta_1=0.5$, and $\beta_2=0.999$. Please refer to our code for more details on the surrogate model's network and training configurations.

\clearpage
\begin{center}
\textbf{\Large Appendix C. Airfoil Design Example: B\'ezierGAN for Smooth Airfoil Generation}
\end{center}
\setcounter{equation}{0}
\setcounter{figure}{0}
\setcounter{table}{0}
\setcounter{section}{0}
\makeatletter
\renewcommand{\theequation}{C.}
\renewcommand{\thefigure}{C.\arabic{figure}}
\renewcommand{\thetable}{C.\arabic{table}}
\renewcommand{\thesection}{C.\arabic{section}}

Samples generated by conventional GANs can be noisy. This noisiness may not be an issue for images, but would cause airfoils to have non-smooth surfaces and greatly impair their performances. Therefore, we replace the conventional GAN with B\'ezierGAN to guarantee the smoothness of generated airfoils. Instead of generating airfoil surface points freely in a 2D space, it uses a B\'ezier layer to constrain generated points to be on a rational B\'ezier curve. We follow \cite{chen2020airfoil} to setup the model architecture. We use the same training data as for the neural network surrogate model. For tests with or without the performance augmented DPP loss (\ie, MO-PaDGAN or GAN), we train the model for 10,000 iterations with a batch size of 32. Both the generator and the discriminator use the Adam optimizer with a learning rate of 0.0001, $\beta_1=0.5$, and $\beta_2=0.999$. Please refer to our code for more details on B\'ezierGAN's network and training configurations.

Note that B\'ezierGAN's generator has both latent codes and the noise vector as inputs. While one can use both inputs as design parameters, we only use the latent codes as variable design parameters in our optimization experiments. How either input affects optimization performance is not central to this work (please refer to \cite{chen2020airfoil} for related details).

\clearpage
\begin{center}
\textbf{\Large Appendix D. Airfoil Design Example: Heuristics for Improving MO-PaDGAN Stability}
\end{center}
\setcounter{equation}{0}
\setcounter{figure}{0}
\setcounter{table}{0}
\setcounter{section}{0}
\makeatletter
\renewcommand{\theequation}{D.\arabic{equation}}
\renewcommand{\thefigure}{D.\arabic{figure}}
\renewcommand{\thetable}{D.\arabic{table}}
\renewcommand{\thesection}{D.\arabic{section}}

While training a MO-PaDGAN, the performances are predicted by $E$ and used in the performance augmented DPP loss to provide feedback for updating the generator. If the quality gradients are not accurate, the generator learning can go astray.
This is not a problem when the quality estimator is a simulator that can reasonably evaluate (even with low-fidelity) any design in the design space, irrespective of the designs being invalid or unrealistic.
However, it creates problems when we use a data-driven surrogate model, as in our airfoil design example. A data-driven surrogate model is normally trained only on realistic designs and hence may perform unreliably on unrealistic ones. In the initial stages of training, a GAN model will not always generate realistic designs. This makes it difficult for the surrogate model to correctly guide the generator's update and may cause stability issues.
We use the follow heuristics to improve the stability:
\begin{enumerate}
\item \textit{Realisticity weighted quality}. We weight the predicted quality at $\mathbf{x}$ by the probability of $\mathbf{x}$ being the real design (predicted by the discriminator):
$$q(\mathbf{x}) = D(\mathbf{x})q'(\mathbf{x})$$
where $q'(\mathbf{x})$ is the predicted quality (by a surrogate model for example), and $D(\mathbf{x})$ is the discriminator's output at $\mathbf{x}$.
\item An \textit{escalating schedule} for setting $\gamma_1$ (the weight of the performance augmented DPP loss). A GAN is more likely to generate unrealistic designs in its early stage of training. Thus, we initialize $\gamma_1$ at 0 and increase it during training, so that MO-PaDGAN focuses on learning to generate realistic designs at the early stage, and takes quality into consideration later when the generator can produce more realistic designs. The schedule is set as:
$$\gamma_1 = \gamma_1'\left(\frac{t}{T}\right)^p$$
where $\gamma_1'$ is the value of $\gamma_1$ at the end of training, $t$ is the current training step, $T$ is the total number of training steps, and $p$ is a factor controlling the steepness of the escalation.
\end{enumerate}

\bibliography{main}

\begin{thebibliography}{10}
\expandafter\ifx\csname url\endcsname\relax
  \def\url#1{\texttt{#1}}\fi
\expandafter\ifx\csname urlprefix\endcsname\relax\def\urlprefix{URL }\fi
\expandafter\ifx\csname href\endcsname\relax
  \def\href#1#2{#2} \def\path#1{#1}\fi

\bibitem{chang2016design}
K.-H. Chang, e-Design: computer-aided engineering design, Academic Press, 2016.

\bibitem{oh2019deep}
S.~Oh, Y.~Jung, S.~Kim, I.~Lee, N.~Kang, Deep generative design: Integration of
  topology optimization and generative models, Journal of Mechanical Design
  141~(11).

\bibitem{burnap2019design}
A.~Burnap, J.~R. Hauser, A.~Timoshenko, Design and evaluation of product
  aesthetics: A human-machine hybrid approach, Available at SSRN 3421771.

\bibitem{shu20203d}
D.~Shu, J.~Cunningham, G.~Stump, S.~W. Miller, M.~A. Yukish, T.~W. Simpson,
  C.~S. Tucker, 3d design using generative adversarial networks and
  physics-based validation, Journal of Mechanical Design 142~(7).

\bibitem{kingma2013auto}
D.~P. Kingma, M.~Welling, Auto-encoding variational bayes, arXiv preprint
  arXiv:1312.6114.

\bibitem{goodfellow2014generative}
I.~Goodfellow, J.~Pouget-Abadie, M.~Mirza, B.~Xu, D.~Warde-Farley, S.~Ozair,
  A.~Courville, Y.~Bengio, Generative adversarial nets, in: Advances in neural
  information processing systems, 2014, pp. 2672--2680.

\bibitem{yang2018microstructural}
Z.~Yang, X.~Li, L.~Catherine~Brinson, A.~N. Choudhary, W.~Chen, A.~Agrawal,
  Microstructural materials design via deep adversarial learning methodology,
  Journal of Mechanical Design 140~(11).

\bibitem{zhang20193d}
W.~Zhang, Z.~Yang, H.~Jiang, S.~Nigam, S.~Yamakawa, T.~Furuhata, K.~Shimada,
  L.~B. Kara, 3d shape synthesis for conceptual design and optimization using
  variational autoencoders, in: ASME 2019 International Design Engineering
  Technical Conferences and Computers and Information in Engineering
  Conference, American Society of Mechanical Engineers Digital Collection,
  2019.

\bibitem{chen2020airfoil}
W.~Chen, K.~Chiu, M.~D. Fuge, Airfoil design parameterization and optimization
  using b{\'e}zier generative adversarial networks, AIAA Journal 58~(11) (2020)
  4723--4735.

\bibitem{chen2020padgan}
W.~Chen, F.~Ahmed, Padgan: Learning to generate high-quality novel designs,
  Journal of Mechanical Design 143~(3).

\bibitem{kulesza2012determinantal}
A.~Kulesza, B.~Taskar, Determinantal point processes for machine learning,
  arXiv preprint arXiv:1207.6083.

\bibitem{salimans2016improved}
T.~Salimans, I.~Goodfellow, W.~Zaremba, V.~Cheung, A.~Radford, X.~Chen,
  Improved techniques for training gans, in: Advances in neural information
  processing systems, 2016, pp. 2234--2242.

\bibitem{chen2021deep}
W.~Chen, A.~Ramamurthy, Deep generative model for efficient 3d airfoil
  parameterization and generation, in: AIAA Scitech 2021 Forum, 2021, p. 1690.

\bibitem{xu2018empirical}
Q.~Xu, G.~Huang, Y.~Yuan, C.~Guo, Y.~Sun, F.~Wu, K.~Weinberger, An empirical
  study on evaluation metrics of generative adversarial networks, arXiv
  preprint arXiv:1806.07755.

\bibitem{couckuyt2014fast}
I.~Couckuyt, D.~Deschrijver, T.~Dhaene, Fast calculation of multiobjective
  probability of improvement and expected improvement criteria for pareto
  optimization, Journal of Global Optimization 60~(3) (2014) 575--594.

\bibitem{mckay2000comparison}
M.~D. McKay, R.~J. Beckman, W.~J. Conover, A comparison of three methods for
  selecting values of input variables in the analysis of output from a computer
  code, Technometrics 42~(1) (2000) 55--61.

\bibitem{rasmussen2006gaussian}
C.~Rasmussen, C.~Williams, M.~Press, F.~Bach, P.~(Firm),
  \href{https://books.google.com/books?id=Tr34DwAAQBAJ}{Gaussian Processes for
  Machine Learning}, Adaptive computation and machine learning, MIT Press,
  2006.
\newline\urlprefix\url{https://books.google.com/books?id=Tr34DwAAQBAJ}

\bibitem{zitzler1999multiobjective}
E.~Zitzler, L.~Thiele, Multiobjective evolutionary algorithms: a comparative
  case study and the strength pareto approach, IEEE transactions on
  Evolutionary Computation 3~(4) (1999) 257--271.

\bibitem{park2010optimal}
K.~Park, J.~Lee, Optimal design of two-dimensional wings in ground effect using
  multi-objective genetic algorithm, Ocean Engineering 37~(10) (2010) 902--912.

\bibitem{drela1989xfoil}
M.~Drela, Xfoil: An analysis and design system for low reynolds number
  airfoils, in: Low Reynolds number aerodynamics, Springer, 1989, pp. 1--12.

\bibitem{he2016deep}
K.~He, X.~Zhang, S.~Ren, J.~Sun, Deep residual learning for image recognition,
  in: Proceedings of the IEEE conference on computer vision and pattern
  recognition, 2016, pp. 770--778.

\bibitem{chen2019aerodynamic}
W.~Chen, K.~Chiu, M.~Fuge, Aerodynamic design optimization and shape
  exploration using generative adversarial networks, in: AIAA SciTech Forum,
  AIAA, San Diego, USA, 2019.

\bibitem{chen2018bezier}
W.~Chen, M.~Fuge, B\'eziergan: Automatic generation of smooth curves from
  interpretable low-dimensional parameters, arXiv preprint arXiv:1808.08871.

\bibitem{chen2016infogan}
X.~Chen, Y.~Duan, R.~Houthooft, J.~Schulman, I.~Sutskever, P.~Abbeel, Infogan:
  Interpretable representation learning by information maximizing generative
  adversarial nets, in: Advances in neural information processing systems,
  2016, pp. 2172--2180.

\bibitem{li2018two}
K.~Li, R.~Chen, G.~Fu, X.~Yao, Two-archive evolutionary algorithm for
  constrained multiobjective optimization, IEEE Transactions on Evolutionary
  Computation 23~(2) (2019) 303--315.
\newblock \href {http://dx.doi.org/10.1109/TEVC.2018.2855411}
  {\path{doi:10.1109/TEVC.2018.2855411}}.

\bibitem{deb2007self}
K.~Deb, K.~Sindhya, T.~Okabe, Self-adaptive simulated binary crossover for
  real-parameter optimization, in: Proceedings of the 9th annual conference on
  genetic and evolutionary computation, 2007, pp. 1187--1194.

\bibitem{deb1996combined}
K.~Deb, M.~Goyal, A combined genetic adaptive search (geneas) for engineering
  design, Computer Science and informatics 26 (1996) 30--45.

\bibitem{poole2015metric}
D.~J. Poole, C.~B. Allen, T.~C. Rendall, Metric-based mathematical derivation
  of efficient airfoil design variables, AIAA Journal 53~(5) (2015) 1349--1361.

\bibitem{poole2019efficient}
D.~J. Poole, C.~B. Allen, T.~Rendall, Efficient aero-structural wing
  optimization using compact aerofoil decomposition, in: AIAA Scitech 2019
  Forum, 2019, p. 1701.

\bibitem{masters2017geometric}
D.~A. Masters, N.~J. Taylor, T.~Rendall, C.~B. Allen, D.~J. Poole, Geometric
  comparison of aerofoil shape parameterization methods, AIAA Journal 55~(5)
  (2017) 1575--1589.

\bibitem{knowles2006parego}
J.~Knowles, Parego: a hybrid algorithm with on-line landscape approximation for
  expensive multiobjective optimization problems, IEEE Transactions on
  Evolutionary Computation 10~(1) (2006) 50--66.

\bibitem{van1999multiobjective}
D.~A. Van~Veldhuizen, G.~B. Lamont, Multiobjective evolutionary algorithm test
  suites, in: Proceedings of the 1999 ACM symposium on Applied computing, 1999,
  pp. 351--357.

\end{thebibliography}

\end{document}